\documentclass[letterpaper, 10 pt, conference]{ieeeconf}
\IEEEoverridecommandlockouts
\usepackage{graphicx,amsmath,amssymb,booktabs,multirow,xcolor,makecell,adjustbox,cite,url,array,capt-of}
\usepackage{placeins,colortbl}
\usepackage{arydshln}
\usepackage{tikz}
\usepackage[hidelinks]{hyperref}

\newcommand{\Qp}{Q_{3\mathrm{P}}}\newcommand{\Qe}{Q_{3\mathrm{E}}}\newcommand{\Qr}{Q_{3\mathrm{R}}}

\newif\ifanon
\anonfalse 

\ifanon
  \hypersetup{pdfauthor={},
    pdfsubject={Preprint under double-blind review at IEEE ICRA 2027}}
\else
  \hypersetup{pdfauthor={Zikai Zhang, Hubert P. H. Shum, Toby P. Breckon},
    pdfsubject={Preprint}}
\fi
\hypersetup{
  pdftitle={VeriDrive: Verifiable Counterfactual Supervision for Budget-Aware Vision-Language Driving},
  pdfkeywords={autonomous driving, vision-language models, counterfactual supervision, graph neural networks},
  pdfcreator={}
}

\title{\LARGE \bf VeriDrive: Verifiable Counterfactual Supervision\\ for Budget-Aware Vision-Language Driving}
\ifanon
  \author{Anonymous Authors%
  \thanks{Preprint. This paper is under double-blind review for IEEE ICRA 2027;
  author names and affiliations are withheld until the review process is
  complete.}}
\else
  \author{Zikai Zhang, Hubert P. H. Shum, and Toby P. Breckon$^{*}$%
  \thanks{The authors are with the Department of Computer Science, Durham
  University, Durham, United Kingdom.}%
  \thanks{$^{*}$Corresponding author.}}
\fi
\IEEEaftertitletext{\begin{minipage}{\textwidth}
\centering
\includegraphics[width=\linewidth]{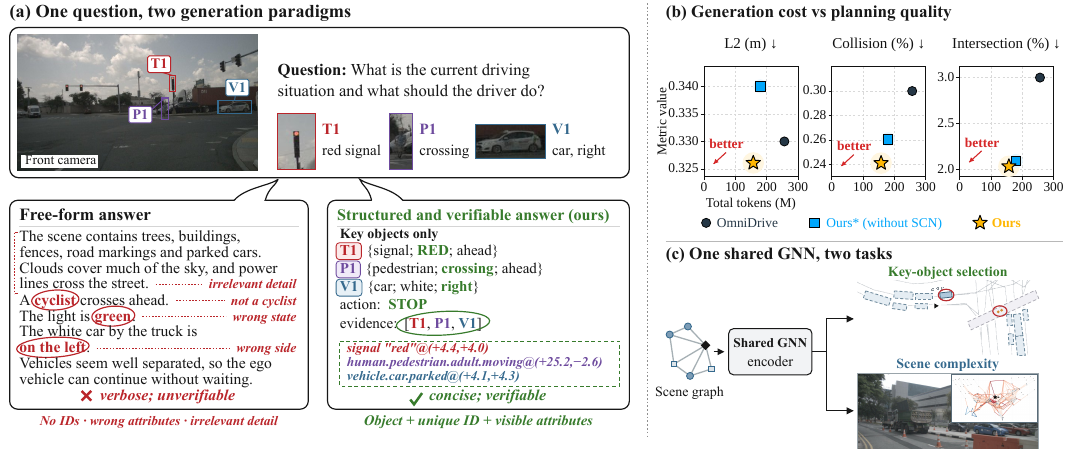}
\captionof{figure}{\textbf{A verifiable and budget-aware framework for VLM-based end-to-end autonomous driving.} (a) The free-form answer contains irrelevant details and incorrect object attributes without explicit evidence links, making the supervision costly and difficult to verify. (b) The nuScenes comparison shows lower token use and improved planning metrics relative to OmniDrive. (c) A shared GNN encoder supports key-object selection and road-scene complexity estimation for budget-aware generation.}
\label{fig:teaser}
\end{minipage}
\vspace{\baselineskip}}
\begin{document}
\makeatletter
\immediate\write\@auxout{\string\citation{veridrive_bst_control}}
\makeatother
\maketitle
\pagestyle{plain}
\providecommand{\codeavail}{[Code will be available after submission].}%
\begin{abstract}
Recent advances in end-to-end learning and vision-language models (VLMs) have substantially improved autonomous driving.
However, VLM-assisted data generation remains susceptible to hallucinations and unverifiable intermediate reasoning, with limited consideration of generation cost and how to allocate VLM resources efficiently.
To address these issues, we propose VeriDrive, a Verifiable and Budget-Aware framework for constructing counterfactual supervision for autonomous driving.
Within a Perception-Evaluation-Reasoning paradigm, VeriDrive constructs compact structured evidence with unique agent and road identifiers and checks spatial relations and counterfactual outcomes against source annotations and explicit rules, making intermediate outputs traceable to source records.
The resulting supervision links grounded perception, rule-checked evaluation, and evidence-based intent revision to the final planning target.
We further introduce the Scene Complexity Network (SCN), which shares a Graph Neural Network (GNN) encoder between key-object selection and road-scene complexity modeling. The two outputs guide evidence filtering and generator selection under a shared budget, with high-quality correction for failed local outputs.
Experiments across multiple planning backbones demonstrate state-of-the-art performance in open-loop settings on nuScenes and closed-loop settings on NAVSIM (nuPlan), while reducing data-generation cost.
\codeavail{}
\end{abstract}

\section{Introduction}
\label{sec:intro}

\noindent 
End-to-end autonomous driving jointly learns perception, prediction, and planning, reducing reliance on separately designed modules~\cite{stp3,uniad,vad}. More recently, vision-language models (VLMs) have introduced semantic knowledge and language reasoning into driving systems~\cite{drivevlm,omnidrive,autovla}. Through intermediate questions and answers, these models connect visual scene understanding to intentions behind planned trajectories~\cite{drivelm}. Counterfactual supervision further exposes the consequences of alternative actions, providing a learning signal beyond imitation of an expert trajectory. Such supervision can teach planners which observed agents and road constraints justify an action, provided its intermediate claims are grounded in the scene.

VLM-generated driving supervision faces reliability and cost challenges. Explanations can refer to nonexistent objects~\cite{pope}, misrepresent spatial relations~\cite{spacedrive}, or revise intentions inappropriately~\cite{cfvla}. CF-VLA~\cite{cfvla} selects scenes for teacher labeling and pairs structured meta-actions with counterfactual reflections, but its reflections lack object- and road-level source identifiers to check individual claims. Manual review assesses consistency, but repeated annotation and regeneration add work~\cite{ye2026autodrivep3}. Long descriptions and irrelevant scene elements also waste tokens. OmniDrive~\cite{omnidrive} grounds counterfactual Q\&A in 3D annotations and rule-based outcome checks, but selects context using fixed proximity cues without learning agent and road relevance. AutoDrive-P$^3$~\cite{ye2026autodrivep3} reduces redundancy through key-object-centered perception-prediction-planning chains, but its reported pipeline does not explicitly budget teacher computation by scene complexity. Generating counterfactual labels for every scene does not consistently improve planning~\cite{cfvla}. This motivates compact, evidence-linked supervision and selective generation that retain the objects and road constraints needed for verification (Fig.~\ref{fig:teaser}).

We propose \textbf{VeriDrive}, a verifiable and budget-aware framework for counterfactual driving supervision. Its Perception-Evaluation-Reasoning (P-E-R) sequence links compact records and scene-scoped agent and road identifiers with rule-checked spatial relations and trajectory outcomes for traceable intent revision and planning. The Scene Complexity Network (SCN) shares a GNN encoder between key-object selection and road-scene complexity heads. Relevance scores filter optional context while retaining mandatory rule witnesses; complexity scores route simple scenes to a local generator and complex scenes to a high-quality generator that also corrects failed local outputs. Across planning backbones, experiments demonstrate state-of-the-art results on nuScenes in open-loop settings and NAVSIM in closed-loop settings, with lower generation cost.

\noindent
Our contributions are threefold:
\begin{itemize}
\item We propose a Verifiable and Budget-Aware framework that constructs compact counterfactual supervision with source-linked evidence, rule-checked intermediate fields, and selective high-quality generation and correction.
\item We introduce SCN with a shared GNN encoder for key-object selection and complex road-scene modeling, using relevance and complexity predictions to guide evidence selection and annotation-budget allocation.
\item We demonstrate effectiveness across planning backbones with state-of-the-art results on nuScenes in open-loop settings and NAVSIM in closed-loop settings, alongside reduced supervision-generation cost.
\end{itemize}

\section{Related Work}
\noindent
We consider prior work across three related sub-topics spanning End-to-End Autonomous Driving (Sec. \ref{sec:related:endtoend}), associated benchmark datasets (Sec. \ref{sec:related:dataset}) and graph representations in the budget-aware space (Sec. \ref{sec:related:gnn}).

\label{sec:related}
\subsection{End-to-End Autonomous Driving}
\label{sec:related:endtoend}
\noindent
ST-P3, UniAD, and VAD integrate perception, prediction, and planning~\cite{stp3,uniad,vad}; LAW and World4Drive add predicted future features~\cite{law,world4drive}. DriveVLM, OmniDrive, and EMMA connect semantic reasoning with planning~\cite{drivevlm,omnidrive,emma}. OpenDriveVLA and ORION introduce spatial interaction modeling and language-conditioned action generation~\cite{opendrivevla,orion}, while SpaceDrive encodes 3D positions~\cite{spacedrive}. AutoVLA combines adaptive reasoning with reinforcement fine-tuning~\cite{autovla}. ImagiDrive refines VLM-based trajectories using imagined futures and limits inference cost through early stopping~\cite{imagidrive}. These planners motivate supervision that supports reliable reasoning without excessive annotation cost.

\subsection{Benchmark QA Driving Datasets}
\label{sec:related:dataset}
\noindent
nuScenes-QA and Reason2Drive provide driving questions and reasoning chains~\cite{nuscenesqa,reason2drive}. DriveLM structures perception, prediction, and planning QA into graphs with object references~\cite{drivelm}. Wang et al.~\cite{ldm_unlabeled} select detector-grounded key objects and refine pseudo-labels using compact scene-graph hints and agreement between generated answers. V2V-GoT links occlusion-aware perception and planning-aware prediction in cooperative-driving graph-of-thoughts supervision~\cite{v2vgot}. OmniDrive grounds counterfactual QA in 3D annotations and trajectory rule checks~\cite{omnidrive}. CF-VLA selects scenes through paired policy rollouts and generates meta-action revisions~\cite{cfvla}, while AutoDrive-P$^3$ combines key-object labels, distance filtering, and manual verification~\cite{ye2026autodrivep3}. These pipelines provide structured supervision and object selection; coupling them with learned agent/road relevance and explicit teacher-budget allocation remains important.

\begin{figure*}[t]
\centering
\includegraphics[width=\textwidth]{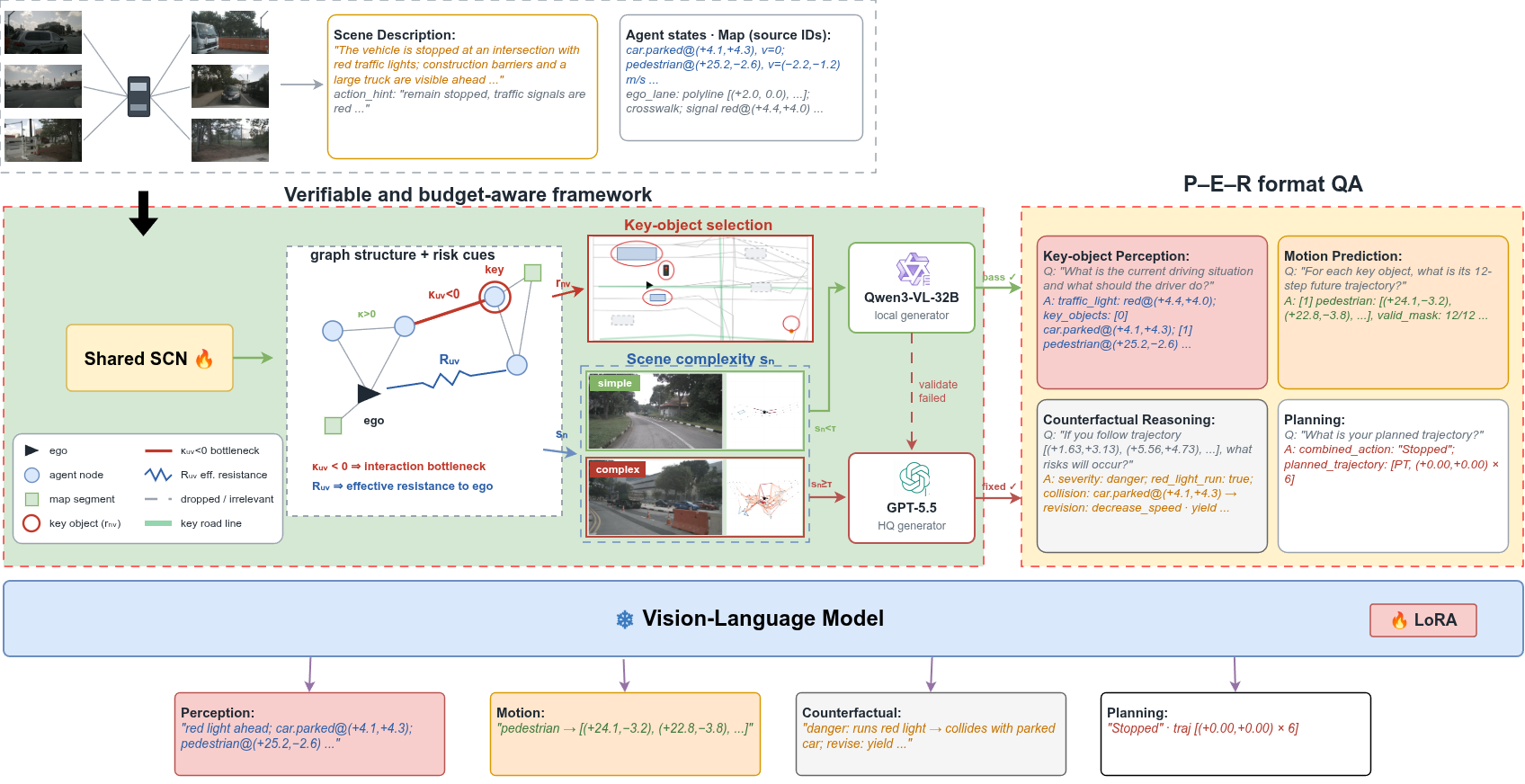}
\caption{Overview of the VeriDrive pipeline. Source-linked records and rule checks provide verifiable counterfactual supervision. A shared GNN encoder supports key-object selection and road-scene complexity modeling for budget-aware generation and correction. The supervision supports multiple planning backbones in open-loop and closed-loop settings.}
\label{fig:framework}
\end{figure*}

\subsection{Graph Representations and Budget-Aware Generation}
\label{sec:related:gnn}
\noindent
HDGT models heterogeneous traffic interactions~\cite{hdgt}; QCNet uses relative spatiotemporal geometry for trajectory prediction~\cite{qcnet}. Logic-RAG converts perceived object relations into first-order logic to improve spatial QA~\cite{logicrag}. Budget allocation complements such grounding: ActiveAD selects clips using planning-oriented diversity and uncertainty~\cite{activead}, and QUORUM routes instances among human and LLM annotators under a fixed budget~\cite{quorum}. AdaThinkDrive learns when chain-of-thought improves planning, adapting inference effort~\cite{adathinkdrive}. For offline teacher allocation, VeriDrive couples checks against scene annotations and trajectory rules with a shared GNN that predicts agent/road relevance and scene complexity.

\section{Methodology}
\label{sec:method}

\noindent 
We offer an initial overview of our approach (Sec. \ref{sec:method:overview}) before detailing verifiable counterfactual supervision (Sec. \ref{sec:supervision}), graph-based scene modeling (Sec. \ref{sec:scn}) and finally our budget-aware generation and correction (Sec. \ref{sec:budget}).

\subsection{Overview}
\label{sec:method:overview}
\noindent
Our VeriDrive approach is summarized in Fig.~\ref{fig:framework}. Given a scene $x_n$ with camera observations, current agent states, and map context, we construct a graph $\mathcal G_n$ and use a shared Scene Complexity Network (SCN) to estimate node relevance $\mathbf r_n$ and scene complexity $s_n$. Relevance scores guide compact evidence selection, while complexity guides the allocation of annotation resources. Recorded agent futures $\mathcal Y_n^*$ and the expert ego trajectory $\tau_n^*$ provide offline targets for counterfactual evaluation and planning. The resulting supervision follows a Perception-Evaluation-Reasoning (P-E-R) sequence:
\begin{equation}
\label{eq:task-interface}
\begin{aligned}
\mathcal G_n&=\Gamma(x_n),\quad (\mathbf r_n,s_n)=f_\theta(\mathcal G_n),\\
\mathcal C_n&=\operatorname{Select}(\mathcal G_n,\mathbf r_n;\mathcal W_n),\\
\mathcal D&=\mathcal A_B\!\left(\{x_n,\mathcal C_n,s_n,\mathcal Y_n^*,\tau_n^*\}_{n=1}^N\right).
\end{aligned}
\end{equation}
Here, $\mathcal W_n$ contains mandatory evidence from full-scene checks, and $\mathcal A_B$ allocates a shared budget $B$ across the scene pool. The complete corpus is $\mathcal D=\{(x_n,\mathcal Q_n)\}_{n=1}^N$, where $\mathcal Q_n$ is the serialized annotation record generated for scene $n$. Its intended schema contains six entries $(Q_1,Q_2,\Qp,\Qe,\Qr,Q_4)$, with scene indices omitted from these field names. Outputs and validation flags are retained even when parsing or consistency checks fail; only passing records are selected for model use.

\subsection{Verifiable Counterfactual Supervision}
\label{sec:supervision}
\begin{figure*}[t]
\centering
\includegraphics[width=0.90\textwidth]{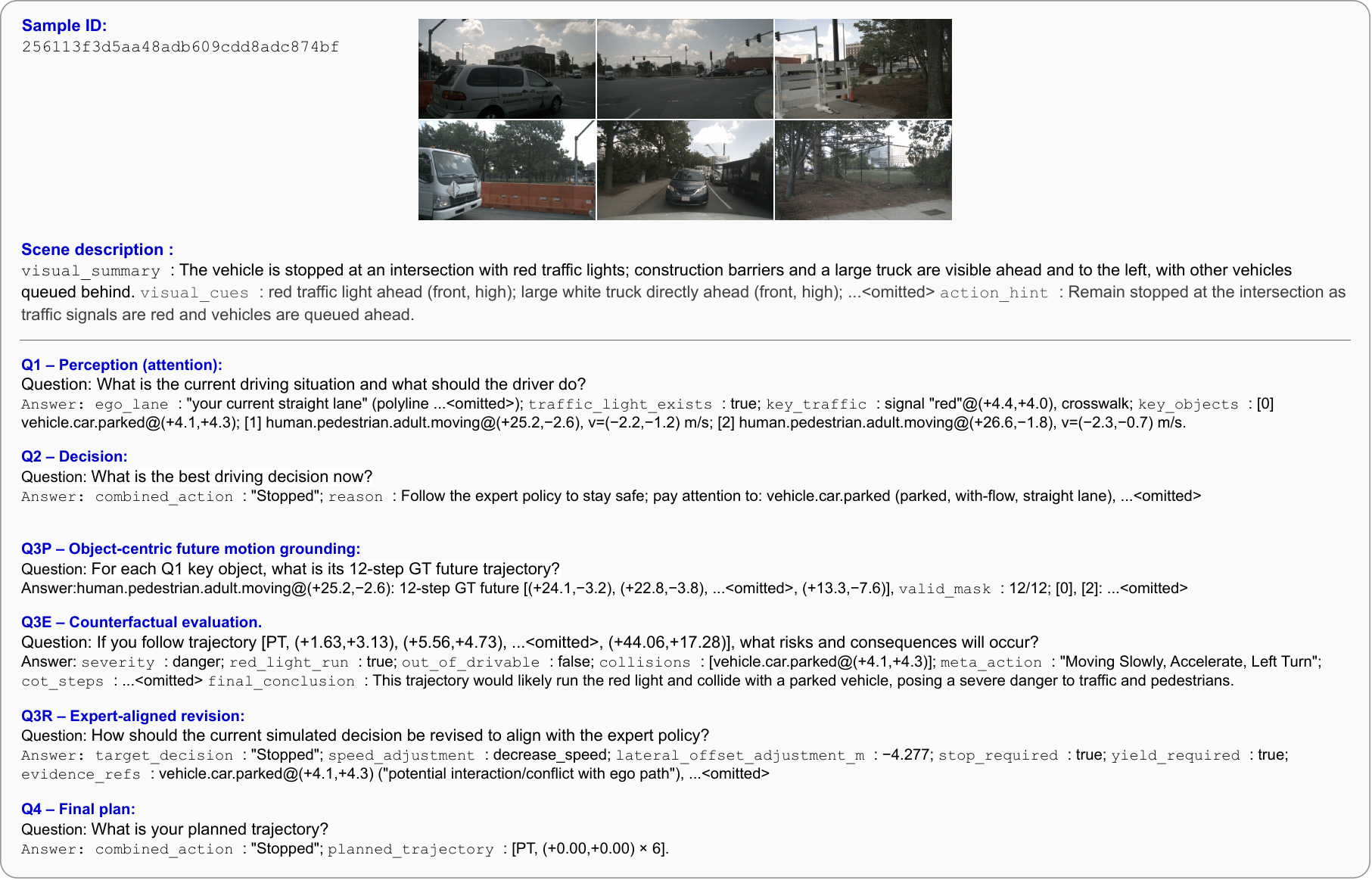}
\vspace{-0.5em}
\caption{An abbreviated example of the generated Perception-Evaluation-Reasoning supervision.}
\label{fig:sample}
\end{figure*}

\noindent
In Fig.~\ref{fig:sample} we illustrate how scene annotations, alternative ego trajectories, and rule-derived outcomes are assembled into supervision. Each record retains scene-scoped source identifiers for agents and road elements, allowing later judgments to refer to the same evidence throughout the sequence.

\textbf{Grounded perception and reasoning.}
$Q_1$ describes selected agents, road geometry, and traffic controls; $Q_2$ gives the expert ego meta-action; and $\Qp$ links object IDs to recorded future positions and validity masks. $\Qr$ revises a counterfactual candidate intent using evidence from $\mathcal C_n$, while $Q_4$ supplies the expert trajectory (Fig.~\ref{fig:sample}).

\textbf{Counterfactual evaluation.}
We construct alternative ego trajectories $\tilde\tau_n^{(m)}$ from motion anchors and connected lane paths with sampled motion profiles~\cite{omnidrive}. Each candidate is evaluated against the complete annotated scene:
\begin{equation}
\label{eq:counterfactual-eval}
\mathcal Z_n^{(m)}=g\!\left(\tilde\tau_n^{(m)};x_n,\mathcal Y_n^*,\mathcal M_n\right),
\end{equation}
where $g$ is a deterministic rule checker, $\mathcal Z_n^{(m)}$ comprises its outcomes and object references, and $\mathcal M_n$ contains map and traffic-control annotations. Collision checks return agent indices whose annotated footprints overlap the candidate ego footprint at valid future steps. Drivable-area checks detect ego occupancy outside the map mask; traffic-light checks test entry into lane regions controlled by red signals using annotated lane-signal associations. Agent futures remain fixed, and $\Qe$ pairs each candidate meta-action with its checked outcomes and evidence. The checker uses the full scene before SCN selection. Mandatory rule witnesses and road constraints $\mathcal W_n$ are retained before optional context; records exceeding the context limit with mandatory evidence alone are deferred for expanded processing.

\vspace{-.25em}
\textbf{Output format and consistency checks.}
A record is eligible for downstream model use only if it parses and agrees with $\mathcal Z_n^{(m)}$: one JSON object with exactly six typed entries covering $Q_1,Q_2,\Qp,\Qe,\Qr,Q_4$, with no undeclared fields, open vocabularies, or malformed coordinates. Object references pair a category token with an ego-frame position, must match an annotated agent of that category within $\epsilon_{\rm ref}=1.0$~m, and must be declared in $Q_1$ when cited by $\Qe$ or $\Qr$. Futures in $\Qp$ are checked against $\mathcal Y_n^*$ under validity masks, and $Q_4$ against $\tau_n^*$ within $\epsilon_{\rm traj}=0.5$~m. Rule fields must reproduce $\mathcal Z_n^{(m)}$ and the revision must match the expert meta-action. Failures are flagged for correction (Sec.~\ref{sec:budget}) and remain in the complete corpus even if unresolved. These checks determine eligibility for model use; explanation faithfulness requires separate semantic auditing.

\subsection{Graph-Based Scene Modeling (SCN)}
\label{sec:scn}
\begin{figure}[t]
\centering
\includegraphics[width=\linewidth]{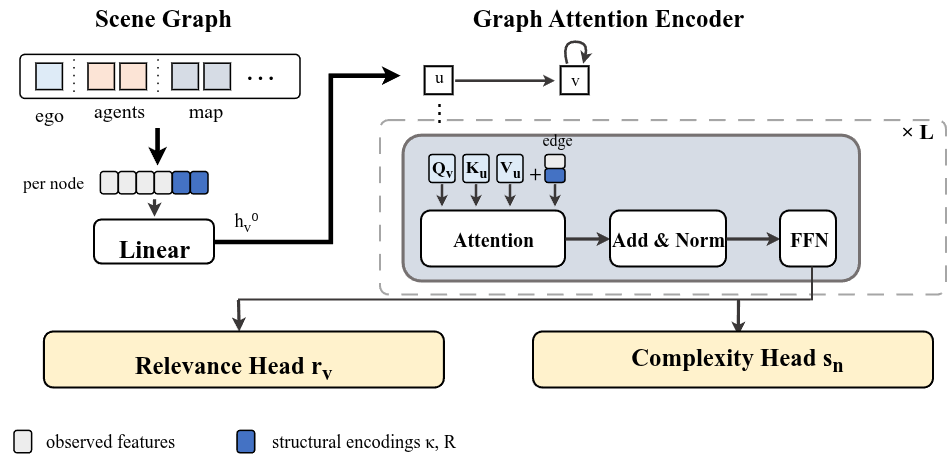}
\caption{Overview of the SCN architecture.}
\label{fig:scn}
\end{figure}

\begin{table*}[t]
\centering\footnotesize\setlength{\tabcolsep}{7.0pt}\renewcommand{\arraystretch}{1.0}
\caption{Open-loop planning results on nuScenes, following the evaluation protocol of OmniDrive~\cite{omnidrive} and ORION~\cite{orion}. Reference results are taken from SpaceDrive~\cite{spacedrive}. Bold marks the best value among the italicized OmniDrive baseline and two VeriDrive variants, rather than the global best.}
\label{tab:nuscenes}
\begin{tabular}{l|cc|ccc>{\columncolor[gray]{0.85}}c|ccc>{\columncolor[gray]{0.85}}c|ccc>{\columncolor[gray]{0.85}}c}
\hline
\multirow{2}{*}{Method} & \multicolumn{2}{c|}{Ego Status} & \multicolumn{4}{c|}{L$_2$ (m) $\downarrow$} & \multicolumn{4}{c|}{Collision (\%) $\downarrow$} & \multicolumn{4}{c}{Intersection (\%) $\downarrow$}\\
 & BEV & Planner & 1s & 2s & 3s & Avg. & 1s & 2s & 3s & Avg. & 1s & 2s & 3s & Avg.\\
\hline
\multicolumn{15}{l}{\textit{Traditional Modular Paradigm}}\\
\hline
ST-P3~\textcolor{blue}{\cite{stp3}} & -- & -- & 1.33 & 2.11 & 2.90 & 2.11 & 0.23 & 0.62 & 1.27 & 0.71 & 2.53 & 8.17 & 14.40 & 8.37\\
UniAD~\textcolor{blue}{\cite{uniad}} & $\checkmark$ & $\checkmark$ & 0.20 & 0.42 & 0.75 & 0.46 & 0.02 & 0.25 & 0.84 & 0.37 & 0.20 & 1.33 & 3.24 & 1.59\\
VAD-Base~\textcolor{blue}{\cite{vad}} & $\checkmark$ & $\checkmark$ & 0.17 & 0.34 & 0.60 & 0.37 & 0.04 & 0.27 & 0.67 & 0.33 & 0.21 & 2.13 & 5.06 & 2.47\\
AD-MLP~\textcolor{blue}{\cite{rethinking_openloop}} & -- & $\checkmark$ & 0.15 & 0.32 & 0.59 & 0.35 & 0.00 & 0.27 & 0.85 & 0.37 & 0.27 & 2.52 & 6.60 & 2.93\\
BEV-Planner++~\textcolor{blue}{\cite{bevplanner}} & $\checkmark$ & $\checkmark$ & 0.16 & 0.32 & 0.57 & 0.35 & 0.00 & 0.29 & 0.73 & 0.34 & 0.35 & 2.62 & 6.51 & 3.16\\
UAD~\textcolor{blue}{\cite{uad}} & $\checkmark$ & $\checkmark$ & 0.13 & 0.28 & 0.48 & 0.30 & 0.00 & 0.19 & 0.61 & 0.27 & 0.13 & 1.08 & 2.89 & 1.37\\
Drive-WM~\textcolor{blue}{\cite{drivewm}} & $\checkmark$ & $\checkmark$ & 0.43 & 0.77 & 1.20 & 0.80 & 0.10 & 0.21 & 0.48 & 0.26 & -- & -- & -- & --\\
\hline
\multicolumn{15}{l}{\textit{VLM-based Paradigm}}\\
\hline
EMMA~\textcolor{blue}{\cite{emma}} & -- & -- & 0.14 & 0.29 & 0.54 & 0.32 & -- & -- & -- & -- & -- & -- & -- & --\\
RDA-Driver~\textcolor{blue}{\cite{rdadriver}} & $\checkmark$ & $\checkmark$ & 0.23 & 0.73 & 1.54 & 0.80 & 0.00 & 0.13 & 0.83 & 0.32 & -- & -- & -- & --\\
DriveVLM~\textcolor{blue}{\cite{drivevlm}} & -- & $\checkmark$ & 0.18 & 0.34 & 0.68 & 0.40 & 0.10 & 0.22 & 0.45 & 0.27 & -- & -- & -- & --\\
ORION~\textcolor{blue}{\cite{orion}} & $\checkmark$ & -- & 0.17 & 0.31 & 0.55 & 0.34 & 0.05 & 0.25 & 0.80 & 0.37 & -- & -- & -- & --\\
SpaceDrive~\textcolor{blue}{\cite{spacedrive}} & -- & -- & 1.06 & 1.79 & 2.55 & 1.80 & 0.35 & 1.33 & 3.97 & 1.88 & 0.96 & 3.38 & 8.28 & 4.21\\
\textit{OmniDrive~\textcolor{blue}{\cite{omnidrive}}} & $\checkmark$ & $\checkmark$ & \textbf{0.14} & \textbf{0.29} & 0.55 & \textbf{0.33} & \textbf{0.00} & 0.13 & 0.78 & 0.30 & 0.56 & 2.48 & 5.96 & 3.00\\
\textit{VeriDrive} & $\checkmark$ & $\checkmark$ & 0.16 & 0.31 & 0.55 & 0.34 & \textbf{0.00} & 0.14 & 0.64 & 0.26 & \textbf{0.53} & 1.88 & 3.87 & 2.09\\
\textit{VeriDrive (+SCN)} & $\checkmark$ & $\checkmark$ & 0.15 & 0.30 & \textbf{0.53} & \textbf{0.33} & \textbf{0.00} & \textbf{0.12} & \textbf{0.61} & \textbf{0.24} & \textbf{0.53} & \textbf{1.82} & \textbf{3.75} & \textbf{2.03}\\
\hline
\multicolumn{15}{l}{\textcolor[gray]{0.5}{\textit{Hybrid Paradigm}}}\\
\hline
\textcolor[gray]{0.5}{VLP~\textcolor{blue}{\cite{vlp}}} & \textcolor[gray]{0.5}{$\checkmark$} & \textcolor[gray]{0.5}{--} & \textcolor[gray]{0.5}{0.30} & \textcolor[gray]{0.5}{0.53} & \textcolor[gray]{0.5}{0.84} & \textcolor[gray]{0.5}{0.55} & \textcolor[gray]{0.5}{0.01} & \textcolor[gray]{0.5}{0.07} & \textcolor[gray]{0.5}{0.38} & \textcolor[gray]{0.5}{0.15} & \textcolor[gray]{0.5}{--} & \textcolor[gray]{0.5}{--} & \textcolor[gray]{0.5}{--} & \textcolor[gray]{0.5}{--}\\
\textcolor[gray]{0.5}{ReAL-AD~\textcolor{blue}{\cite{realad}}} & \textcolor[gray]{0.5}{$\checkmark$} & \textcolor[gray]{0.5}{--} & \textcolor[gray]{0.5}{0.30} & \textcolor[gray]{0.5}{0.48} & \textcolor[gray]{0.5}{0.67} & \textcolor[gray]{0.5}{0.48} & \textcolor[gray]{0.5}{0.07} & \textcolor[gray]{0.5}{0.10} & \textcolor[gray]{0.5}{0.28} & \textcolor[gray]{0.5}{0.15} & \textcolor[gray]{0.5}{--} & \textcolor[gray]{0.5}{--} & \textcolor[gray]{0.5}{--} & \textcolor[gray]{0.5}{--}\\
\textcolor[gray]{0.5}{DriveVLM-Dual~\textcolor{blue}{\cite{drivevlm}}} & \textcolor[gray]{0.5}{$\checkmark$} & \textcolor[gray]{0.5}{--} & \textcolor[gray]{0.5}{0.15} & \textcolor[gray]{0.5}{0.29} & \textcolor[gray]{0.5}{0.48} & \textcolor[gray]{0.5}{0.31} & \textcolor[gray]{0.5}{0.05} & \textcolor[gray]{0.5}{0.08} & \textcolor[gray]{0.5}{0.17} & \textcolor[gray]{0.5}{0.10} & \textcolor[gray]{0.5}{--} & \textcolor[gray]{0.5}{--} & \textcolor[gray]{0.5}{--} & \textcolor[gray]{0.5}{--}\\
\textcolor[gray]{0.5}{SOLVE-VLM~\textcolor{blue}{\cite{solve}}} & \textcolor[gray]{0.5}{$\checkmark$} & \textcolor[gray]{0.5}{--} & \textcolor[gray]{0.5}{0.13} & \textcolor[gray]{0.5}{0.25} & \textcolor[gray]{0.5}{0.47} & \textcolor[gray]{0.5}{0.28} & \textcolor[gray]{0.5}{0.00} & \textcolor[gray]{0.5}{0.16} & \textcolor[gray]{0.5}{0.43} & \textcolor[gray]{0.5}{0.20} & \textcolor[gray]{0.5}{--} & \textcolor[gray]{0.5}{--} & \textcolor[gray]{0.5}{--} & \textcolor[gray]{0.5}{--}\\
\textcolor[gray]{0.5}{Senna~\textcolor{blue}{\cite{senna}}} & \textcolor[gray]{0.5}{$\checkmark$} & \textcolor[gray]{0.5}{--} & \textcolor[gray]{0.5}{0.11} & \textcolor[gray]{0.5}{0.21} & \textcolor[gray]{0.5}{0.35} & \textcolor[gray]{0.5}{0.22} & \textcolor[gray]{0.5}{0.04} & \textcolor[gray]{0.5}{0.08} & \textcolor[gray]{0.5}{0.13} & \textcolor[gray]{0.5}{0.08} & \textcolor[gray]{0.5}{--} & \textcolor[gray]{0.5}{--} & \textcolor[gray]{0.5}{--} & \textcolor[gray]{0.5}{--}\\
\hline
\end{tabular}
\end{table*}

\begin{table}[t]
\centering\footnotesize\setlength{\tabcolsep}{1.8pt}\renewcommand{\arraystretch}{1.0}
\caption{NAVSIM closed-loop settings. Higher is better. Bold: global best, including ties. Underline: best below the dashed line.}
\label{tab:navsim}
\begin{tabular}{lrrrrrrr}
\toprule
Method & PDMS & Coll. & Area & Dir. & Prog. & TTC & Comf.\\
\midrule
Ego Status MLP~\textcolor{blue}{\cite{navsim}} & 66.40 & 93.09 & 78.26 & 90.45 & 63.20 & 84.02 & 99.97\\
TransFuser~\textcolor{blue}{\cite{transfuser}} & 83.88 & 97.78 & 92.63 & 97.97 & 78.88 & 92.89 & 99.98\\
DRAMA~\textcolor{blue}{\cite{drama_navsim}} & 86.87 & 98.19 & 95.18 & \textbf{98.03} & 81.33 & 94.17 & \textbf{100.00}\\
Hydra-MDP~\textcolor{blue}{\cite{hydramdp}} & 91.26 & 99.07 & 98.29 & 95.79 & 85.20 & 96.56 & \textbf{100.00}\\
Centaur~\textcolor{blue}{\cite{centaur}} & 92.10 & 99.23 & \textbf{98.72} & 96.77 & 85.96 & 97.17 & 99.97\\
TrajHF~\textcolor{blue}{\cite{trajhf}} & \textbf{93.95} & 99.30 & 97.51 & 91.72 & \textbf{90.39} & 98.02 & 99.81\\
\hdashline
\textit{AutoVLA~\textcolor{blue}{\cite{autovla}}} & 92.12 & 99.14 & 97.08 & 95.51 & 87.55 & 97.12 & \underline{99.98}\\
\textit{VeriDrive} & \underline{93.40} & \textbf{\underline{99.42}} & \underline{97.97} & \underline{96.61} & \underline{88.63} & \textbf{\underline{98.11}} & 99.97\\
\hline
\end{tabular}
\end{table}

\begin{figure*}[t]
\centering
\includegraphics[width=0.70\linewidth]{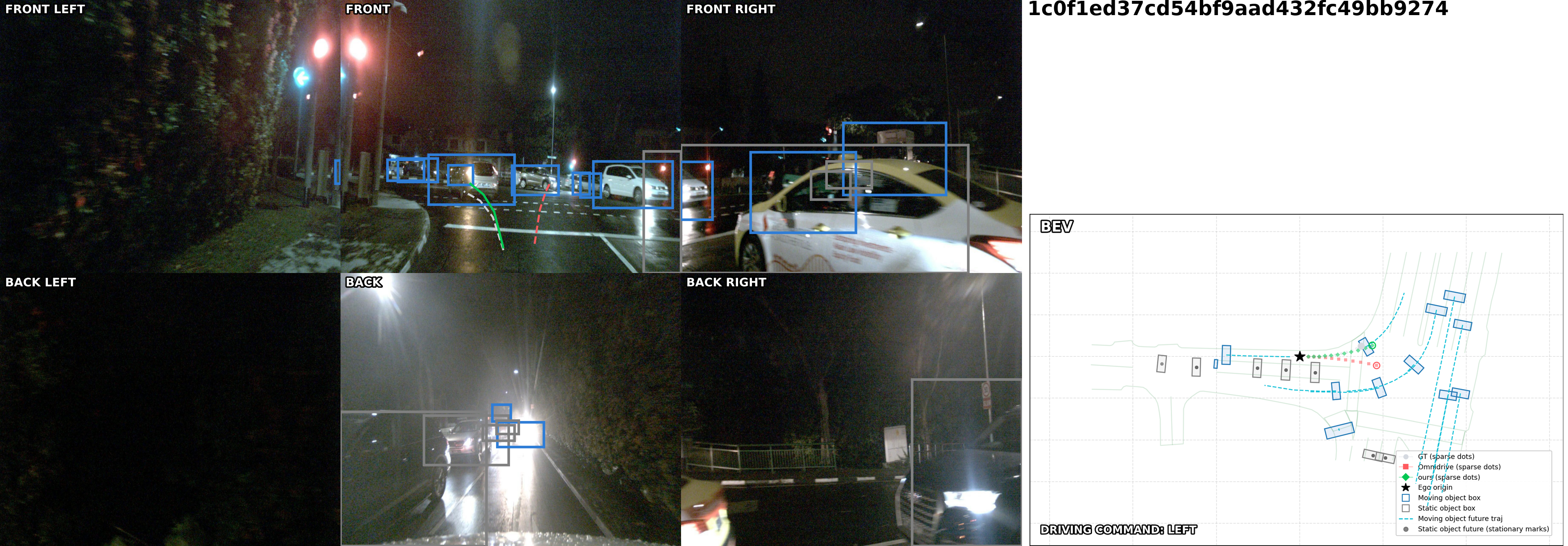}
\vspace{0.2em}
\includegraphics[width=0.70\linewidth]{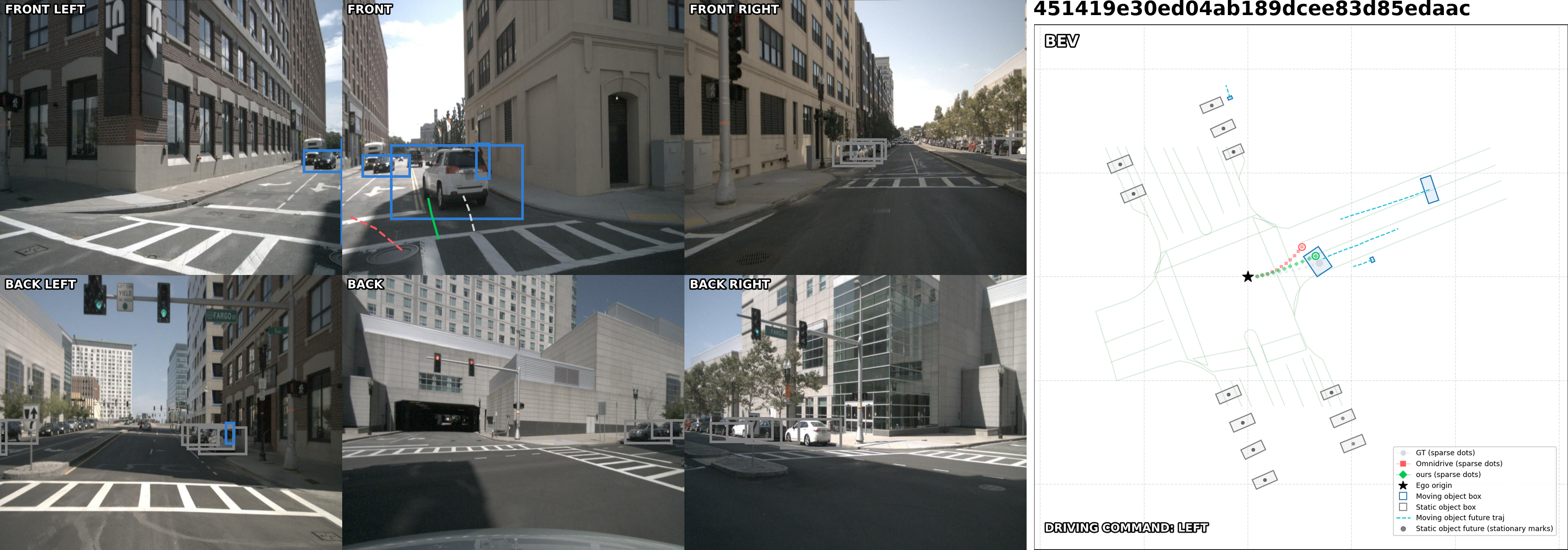}
\vspace{-0.3em}
\caption{Qualitative planning comparison on two intersection scenes. Left: multi-view context. Right: BEV trajectories, showing the \textcolor{gray}{expert ($\bullet$)}, \textcolor{red}{OmniDrive ($\blacksquare$)} and \textcolor{green!60!black}{VeriDrive ($\blacklozenge$)} paths from the ego origin ($\bigstar$), with \textcolor{blue}{moving} and \textcolor{gray}{static} object boxes and \textcolor{cyan!70!black}{dashed object futures}. Upper: night intersection. Lower: daytime left turn. The nuScenes sample token is printed above each BEV.\vspace{-1em}}
\label{fig:qualitative}
\vspace{-1em}
\end{figure*}

\noindent
SCN shares one graph encoder between evidence selection and complexity estimation (Fig.~\ref{fig:scn}). Its single-frame graph $\mathcal G_n=(V,E)$ contains ego, agent, and map-segment nodes. Radius-limited nearest-neighbor edges connect agents; proximity connects agents to map segments; adjacency and nearby endpoints connect segments. Node features encode category, pose, velocity, and size. Typed edge features $\mathbf d_{uv}$ encode relative distance, bearing, heading, and velocity in the receiving node frame.

\textbf{Curvature and resistance encodings.}
We compute structural features on the unweighted, undirected simple projection of $\mathcal G_n$. Balanced Forman curvature~\cite{topping2022curvature} characterizes the local connectivity around an edge. For endpoint degrees $d_u,d_v>1$, it is
\begin{equation}
\label{eq:curvature}
\begin{aligned}
\kappa_{uv}={}&\frac{2}{d_u}+\frac{2}{d_v}-2
 +\frac{2t_{uv}}{d_{\max}}+\frac{t_{uv}}{d_{\min}}\\
&+\frac{s_{u|v}+s_{v|u}}{\gamma_{uv}d_{\max}},
\end{aligned}
\end{equation}
where $d_{\min},d_{\max}$ are the endpoint degree extrema, $t_{uv}$ counts shared neighbors, and $s_{u|v},s_{v|u}$ count distinct neighbors on either side participating in chordless four-cycles through the edge. $\gamma_{uv}$ is the largest number of such cycles sharing one of these neighbors. The last term is zero when no such cycle exists; $\kappa_{uv}=0$ if $d_{\min}\le1$. Negative curvature provides a cue to structural bottlenecks.

Effective resistance captures connectivity through the wider graph~\cite{black2023resistance}. For nodes in the same connected component $C$, with unit edge conductance,
\begin{equation}
\label{eq:resistance}
R_{uv}=(\mathbf e_u-\mathbf e_v)^\top L_C^\dagger(\mathbf e_u-\mathbf e_v),
\quad L_C=D_C-A_C,
\end{equation}
where $\mathbf e_v$ is a basis vector and $\dagger$ denotes the pseudoinverse. We augment each node with its local curvature profile~\cite{lcp} (minimum, mean, maximum, standard deviation, and median of incident curvatures), resistance curvature $p_v=1-\frac12\sum_{u\sim v}R_{uv}$~\cite{devriendt2022resistance}, and resistance to ego. These features are standardized within each graph; ego-unreachable nodes receive a neutral zero after exclusion from ego-resistance normalization. The network learns how these structural features contribute to relevance and complexity predictions.

\textbf{Shared graph attention.}
A linear projection of node and structural features initializes $\mathbf h_v^0$. For edge $u\!\to\!v$ of type $t$, each attention head uses
\begin{equation}
\label{eq:scn-attention}
\begin{aligned}
e_{vu}&=\frac{(W_Q\mathbf h_v)^\top(W_K\mathbf h_u)}{\sqrt{d_h}}
       +\mathbf w_t^\top\mathbf d_{uv}+b_t\\
&\phantom{=}+\alpha_t\widetilde\kappa_{uv}+\beta_t\widetilde R_{uv},
\end{aligned}
\end{equation}
where tildes denote graph-wise standardization and $d_h$ is the head dimension. Head and layer indices are omitted for clarity. The learned, type-specific bias parameters are shared across layers. Softmax over incoming neighbors and a self-loop weights the value projections, followed by multihead aggregation, residual normalization, and a feed-forward block.

\textbf{Relevance and complexity heads.}
The final node embeddings predict relevance independently, while ego-conditioned attention pools all nodes for scene classification:
\begin{equation}
\label{eq:scn-heads}
\begin{aligned}
\pi_v&=\operatorname{softmax}_v
 \frac{(W_K^p\mathbf h_v)^\top(W_Q^p\mathbf h_{\rm ego})}{\sqrt d},\\
\mathbf h_n^{\rm pool}&=\sum_v\pi_v\mathbf h_v,\quad
r_v=\sigma(\operatorname{MLP}_r(\mathbf h_v)),\\
s_n&=\sigma(\operatorname{MLP}_s([\mathbf h_{\rm ego},\mathbf h_n^{\rm pool},\boldsymbol\zeta_n])).
\end{aligned}
\end{equation}
Here $d$ is the embedding dimension and $\mathbf h_n^{\rm pool}\in\mathbb R^d$ is the ego-conditioned pooled scene embedding. The graph statistics $\boldsymbol\zeta_n$ retain absolute structural information removed by standardization: mean and minimum curvature, negative-edge fraction, $\log(1+R_{\rm tot}/|V|)$, mean edge resistance, and $\log(1+|V|),\log(1+|E|)$. $R_{\rm tot}$ sums resistance over unordered within-component node pairs; edge statistics use the original directed edges before self-loops. The two heads are jointly optimized with
\begin{equation}
\label{eq:scn-loss}
\mathcal L=\sum_n w_n\,\ell(s_n,q_n)
+\lambda\sum_{n,v}m_{nv}\,\ell(r_{nv},z_{nv}),
\end{equation}
where $\ell$ is binary cross-entropy, $q_n$ and $z_{nv}$ are complexity and relevance targets, $w_n$ weights scene supervision, and $m_{nv}$ masks unavailable node targets. The outputs guide context selection and annotation allocation, respectively.

\subsection{Budget-Aware Generation and Correction}
\label{sec:budget}
As shown in Fig.~\ref{fig:framework}, SCN routes simple scenes ($s_n<\eta$) to a local open-weight VLM and complex scenes ($s_n\ge\eta$) directly to a high-quality VLM, where $\eta\in[0,1]$ is the complexity-score threshold. Both generators receive the selected evidence $\mathcal C_n$ and rule outcomes. Local outputs that fail validation are sent to the high-quality model with checker feedback for correction. The threshold is set under the shared budget $B$, with a portion reserved for these corrections. Outputs from both branches undergo the same checks: passing records are selected for downstream model use, while all outputs remain in the complete corpus for quality measurement. We report token usage, wall-clock time, and estimated API cost for generation and correction (Table~\ref{tab:gencost}).

\vspace{-0.1cm}
\section{Experiments}
\label{sec:experiments}
\vspace{-0.15cm}
\subsection{Experimental Settings}
\vspace{-0.15cm}
\noindent
\textbf{Datasets and metrics.}
We evaluate open-loop planning on nuScenes~\cite{nuscenes} and planning in closed-loop settings on NAVSIM~\cite{navsim}. On nuScenes we report displacement error (L$_2$), collision rate, and drivable-area intersection rate at 1, 2 and 3~s with their averages; lower is better. As in AutoVLA~\cite{autovla}, NAVSIM queries the planner once and simulates its fixed plan against logged, non-reactive traffic, without replanning or environmental feedback. On NAVSIM we report the predictive driver model score (PDMS) together with its no-collision, drivable-area, direction, progress, time-to-collision and comfort terms. Supervision is generated on the nuScenes training split from its 3D annotations and the OpenLane-V2 map and traffic-control labels~\cite{openlanev2}. After discarding frames without a valid expert trajectory, the corpus holds 27{,}271 training and 5{,}868 validation records.

\noindent
\textbf{Implementation details.}
We use Qwen3-VL-32B~\cite{qwen3vl} for local generation and GPT-5.5 for high-quality generation and correction. Unless stated otherwise, we retain at most $K=5$ key objects and two traffic elements per record, use $T=1$ counterfactual candidate per scene, and use a $30\%$ high-quality routing ratio. Planner fine-tuning runs on four NVIDIA A100 GPUs. We adopt the Omni-Q planner of OmniDrive~\cite{omnidrive} for nuScenes and AutoVLA~\cite{autovla} for NAVSIM, replacing their final-stage supervision with our P-E-R records. All other training and evaluation settings follow the respective baselines. Recorded futures and rule outcomes are used only for offline supervision; at inference, each planner receives its baseline observations.

\subsection{Quantitative Results}

\noindent
\textbf{Open-loop planning.}
Table~\ref{tab:nuscenes} compares open-loop planning results. OmniDrive~\cite{omnidrive} is the baseline with the same backbone and ego-status setting. VeriDrive (+SCN) lowers the average collision rate from $0.30$ to $0.24\%$ and the average intersection rate from $3.00$ to $2.03\%$ at a comparable average L$_2$ of $0.326$~m; the margins are largest at the 3~s horizon. Selecting relevant agents and road elements improves all three averages over VeriDrive without SCN. Other published baselines use different backbones, inputs or ego-status conventions.

\noindent
\textbf{Closed-loop settings.}
On NAVSIM's non-reactive evaluation (Table~\ref{tab:navsim}), VeriDrive achieves a PDMS of $93.40$, exceeding the published AutoVLA~\cite{autovla} reference of $92.12$ by $1.28$ points. AutoVLA's reported result uses oracle selection from six generated trajectories. VeriDrive scores higher in no-collision ($99.42$ vs.\ $99.14$), drivable-area compliance, driving direction, progress, and TTC ($98.11$ vs.\ $97.12$), with slightly lower comfort ($99.97$ vs.\ $99.98$).

\subsection{Qualitative Results}
Fig.~\ref{fig:qualitative} compares the two policies on intersection scenes under identical context. In the night case (upper), the baseline drifts laterally from the expert path while VeriDrive stays close to it across the full horizon. In the daytime scene (lower), VeriDrive remains closer to the expert trajectory, while the baseline deviates to its left.

\subsection{Ablation Studies}
\begin{table}[t]
\centering\footnotesize\setlength{\tabcolsep}{3.2pt}\renewcommand{\arraystretch}{1.0}
\caption{Ablation studies on nuScenes: high-quality routing, intermediate supervision, and key-object and counterfactual counts.}
\label{tab:ablation}
\begin{tabular}{@{}l r ccc@{}}
\toprule
Setting & HQ (\%) & L$_2$ (m) & Coll. (\%) & Inter. (\%)\\
\midrule
\multicolumn{5}{@{}l}{\textit{High-quality routing}}\\
Local only & 0 & 0.4478 & 0.3749 & 2.1764\\
Random correction & 25 & 0.3942 & 0.3396 & 2.1398\\
Validator routing & 25 & 0.3294 & 0.2468 & 2.0417\\
Random correction & 50 & 0.3647 & 0.3062 & 2.1065\\
Validator routing & 50 & 0.3265 & 0.2414 & 2.0344\\
All high-quality & 100 & 0.3268 & \textbf{0.2161} & 2.0573\\
VeriDrive (+SCN) & 30 & \textbf{0.3261} & 0.2411 & \textbf{2.0328}\\
\midrule
\multicolumn{5}{@{}l}{\textit{Progressive intermediate supervision}}\\
$Q_4$ only & -- & 0.3912 & 0.3602 & 2.3569\\
+ $\Qp$ future grounding & -- & 0.3834 & 0.3187 & 2.2974\\
+ $\Qe$ evaluation & -- & 0.3470 & 0.3048 & 2.0980\\
+ $\Qr$ revision & -- & \textbf{0.3261} & \textbf{0.2411} & \textbf{2.0328}\\
\midrule
\multicolumn{5}{@{}l}{\textit{Key objects $K$ and counterfactual candidates $T$}}\\
$K=3$ & -- & 0.5896 & 0.4138 & 2.4517\\
$K=5$ & -- & \textbf{0.3261} & \textbf{0.2411} & \textbf{2.0328}\\
$K=8$ & -- & 0.3389 & 0.2597 & 2.0794\\
\midrule
$T=1$ & -- & \textbf{0.3261} & \textbf{0.2411} & \textbf{2.0328}\\
$T=2$ & -- & 0.3354 & 0.2575 & 2.0593\\
$T=4$ & -- & 0.3336 & 0.2558 & 2.0478\\
\bottomrule
\end{tabular}
\end{table}

\begin{table}[t]\centering\footnotesize\setlength{\tabcolsep}{3pt}\renewcommand{\arraystretch}{1.0}
\caption{Verification diagnostics on the complete corpus after generation and correction, before selection for model use (D), and no-oracle rollouts (I).}
\label{tab:verification}
\begin{tabular}{@{}lcr@{}}
\toprule
Diagnostic & Stage & Pass (\%)\\
\midrule
Format validity & D & 98.7\\
Required-field completeness & D & 97.9\\
Object-ID validity & D & 99.1\\
$\Qe$ rule consistency & D & 96.4\\
$\Qr$ expert-action agreement & D & 91.2\\
Full-chain parse validity & I & 100.0\\
$\hat{Q}_{3\mathrm{P}}$ object-ID validity & I & 95.5\\
$\hat{Q}_{3\mathrm{E}}$ checker consistency & I & 100.0\\
$\hat{Q}_4$ trajectory parse validity & I & 99.0\\
\bottomrule
\vspace{-3em}
\end{tabular}
\end{table}
\begin{table}[t]\centering\footnotesize\setlength{\tabcolsep}{2.6pt}\renewcommand{\arraystretch}{1.0}
\caption{Data-generation cost comparison: high-quality routing ratio, token usage, generation time, and API cost.}
\label{tab:gencost}
\begin{tabular}{@{}lrrrrr@{}}
\toprule
Pipeline & HQ (\%) & Tok./sample & Tok. (M) & Time (h) & API (\$k)\\
\midrule
OmniDrive~\cite{omnidrive} & 100 & 7524.3 & 256.6 & 290.0 & 2.69\\
VeriDrive & 30 & 5421.1 & 179.6 & 238.0 & 0.57\\
VeriDrive (+SCN) & 30 & \textbf{4746.2} & \textbf{157.3} & \textbf{208.2} & \textbf{0.54}\\
\bottomrule
\vspace{-3em}
\end{tabular}
\end{table}

\noindent
\textbf{Routing strategies.}
The upper panel of Table~\ref{tab:ablation} compares routing policies at several high-quality usage ratios. The local-only configuration has the highest errors. At matched $25\%$ and $50\%$ ratios, validator routing gives lower average L$_2$ than random routing ($0.329$ vs.\ $0.394$ and $0.327$ vs.\ $0.365$, respectively). At the $30\%$ high-quality routing setting, VeriDrive (+SCN) has displacement and intersection errors close to all-high-quality generation, which has the lowest collision rate.

\noindent
\textbf{Core components.}
The middle panel evaluates cumulative supervision stages. Adding rule-checked evaluation $\Qe$ gives the largest L$_2$ reduction ($0.3834$ to $0.3470$~m), while expert-aligned revision $\Qr$ gives the largest collision-rate reduction ($0.3048$ to $0.2411\%$). The complete sequence has the lowest errors in all three metrics. Among the key-object counts and counterfactual candidate counts tested in the lower panel, $K=5$ and $T=1$ give the lowest errors.

\noindent
\textbf{Supervision quality.}
Table~\ref{tab:verification} reports diagnostics on the complete corpus after generation and correction, including records that fail validation. The validator selects records for model use without deleting them from this corpus. Format validity, required-field completeness and object-ID validity all exceed $97\%$, $\Qe$ agrees with the rule checker on $96.4\%$ of records, and $\Qr$ matches the expert meta-action on $91.2\%$. In no-oracle rollouts the fine-tuned planner emits a parseable chain on every scene and a parseable trajectory on $99.0\%$. These diagnostics cover specific fields and do not establish explanation faithfulness.

\noindent
\textbf{Generation cost.}
Table~\ref{tab:gencost} compares the reported generation costs. VeriDrive (+SCN) uses $37\%$ fewer tokens per sample than OmniDrive. Project generation time decreases from $290.0$ to $208.2$~h and estimated API cost from \$2.69k to \$0.54k. These project totals cover approximately 34.1k samples for OmniDrive and 33.1k for VeriDrive. They measure VLM generation and correction; SCN preprocessing, labeling, and training overhead are not included.

\vspace{-0.3em}
\section{Conclusion}
\vspace{-0.2em}
\label{sec:conclusion}
\noindent
We presented VeriDrive, a framework for verifiable and budget-aware counterfactual driving supervision. Scene-scoped identifiers and rule checks link intermediate outputs to source annotations, while a shared graph encoder selects relevant evidence and estimates scene complexity to guide generation and correction. On nuScenes, the $30\%$ high-quality routing setting gives displacement and intersection errors comparable to all-high-quality generation. The reported API cost is about one fifth of OmniDrive's. Explanation faithfulness and SCN-guided annotation in reactive closed-loop driving require further evaluation.

\FloatBarrier
\nocite{spacedrive,drivevlm,nuscenes,openlanev2,bevplanner,law,world4drive,solve,opendrivevla,navsim,lcp,black2023resistance,reason2drive,nuscenesqa,qwen3vl}
\vspace{-.35em}
\bibliographystyle{IEEEtran}
\bibliography{refs}
\end{document}